# Feature Selection in Conditional Random Fields for Map Matching of GPS Trajectories

Jian Yang, Liqiu Meng

Lehrstuhl für Kartographie, Technische Universität München, 80333 Munich, Germany

**Abstract.** Map matching of the GPS trajectory serves the purpose of recovering the original route on a road network from a sequence of noisy GPS observations. It is a fundamental technique to many Location Based Services. However, map matching of a low sampling rate on urban road network is still a challenging task. In this paper, the characteristics of Conditional Random Fields with regard to inducing many contextual features and feature selection are explored for the map matching of the GPS trajectories at a low sampling rate. Experiments on a taxi trajectory dataset show that our method may achieve competitive results along with the success of reducing model complexity for computation-limited applications.

## 1. Introduction

Map matching of GPS trajectory serves the purpose of recovering the original route on a road network from a sequence of GPS observations. It is a fundamental technique for many Location Based Services (LBS) as it brings added value to the raw GPS data and has the potential to distill more reliable knowledge about routing on road networks. However, the GPS observations are often noisy so that finding the nearest roads usually fails. Many research works have been dedicated to map matching of GPS trajectory with a moderate sampling rate, while map matching with a low sampling rate, namely the sampling interval greater than 120 seconds, is still an ongoing research topic in recent years (Hunter et al., 2013; Li et al., 2013).

Map matching is often modeled as a sequence labeling problem. The Hidden Markov Model (HMM) and its variants have been intensively explored in previous attempts (Hummel, 2006; Krumm et al., 2007; Lou et al., 2009; Newson & Krumm, 2009; Yuan et al., 2010). Being constrained by the strict statistical assumptions, however, these generative models fail to capture the non-independent characteristics from sparse GPS observations and therefore result in poor performance. This gives rise to Conditional Random Fields (CRFs) (Lafferty et al., 2001), another probabilistic model for labeling sequential data that allows to use many non-independent and overlapped features drawn from observations to improve the matching accuracy. However, the CRFs requires intensive computation which could prohibit computation-limited applications such as the map matching on mobile devices. This constraint stimulates the need to select the most relevant feature subset in the CRFs, thus reduce the model complexity in terms of the number of features.

In this paper, we attempt to construct a compact CRFs for map matching through feature selection. More specifically, we first induce rich features to CRFs for map matching, and then train the CRFs with $\ell_1$ regularization to yield a sparse model (many features are assigned to zero weights). To verify the effectiveness of feature selection, we perform an experiment on a sample dataset derived from Taxi Floating Car Data (FCD) in Shanghai, China. Following contributions could be highlighted in our work:

1. We explore the further use of the CRFs for map matching to yield a sparse model with a higher matching accuracy via feature selection. Experiment shows that 50% feature reduction and 10% accuracy improvement can be achieved compared to a common model.
2. As we induce features from most cited literatures, the result of feature selection can serve as an experimental review of previous modeling effort and provide guidance for designing simple model for map matching.
3. The learned weights of selected features also reflect road usage pattern in the study area.

## 2. Map Matching of GPS Trajectory

Map matching requires both GPS observations and a road network. The basic attributes of the observations collected by positioning sensors include latitude, longitude and timestamp, while extra information such as instant speed, acceleration, heading direction etc. can also be obtained from the sensors. Due to the inaccuracy embedded in both observations and the road network, the matching of the nearest road matching often fails and it is therefore necessary to develop map matching methods. The challenge of this task is two-folded: 1) Observations are often noisy due to inaccurate GPS sensor or poor positioning conditions, e.g. low-speed maneuvers of vehicle in traffic, passage through urban canyon and tunnels. These facts make map matching problematic in a dense road network. 2) A low sampling rate which aims to reduce the communication cost or data storage causes information loss between neighboring observations, making the route recovery extremely difficult as a huge number of feasible paths can be found on road network. An example of GPS trajectory is illustrated in *Figure 1*.

Map matching has invoked a growing interest in the past years for its importance in LBS applications. A comprehensive literature survey was done in (Quddus et al., 2007), in which map matching method is categorized into four groups: geometric, topological, probabilistic and other advanced techniques. Among these approaches, the Hidden Markov Model (HMM)-based probabilistic methods are most popular because of their well-studied theoretical base and competitive performances. A HMM-based method models the probability of a sequence possible road assignments on road network for given GPS observations. The computation of the sequence probability requires a strict modeling of the observation probability and transition probability, namely probability for candidate roads for GPS observations and candidate paths in between. These two components are designed to capture the characteristics of noisy sensors and the original route from different perspectives. And matching GPS trajectories at a low sampling rate often requires richer features of observation and transition for better accuracy. However, this would cause an intractable inference problem for HMM models (Lafferty et al., 2001).

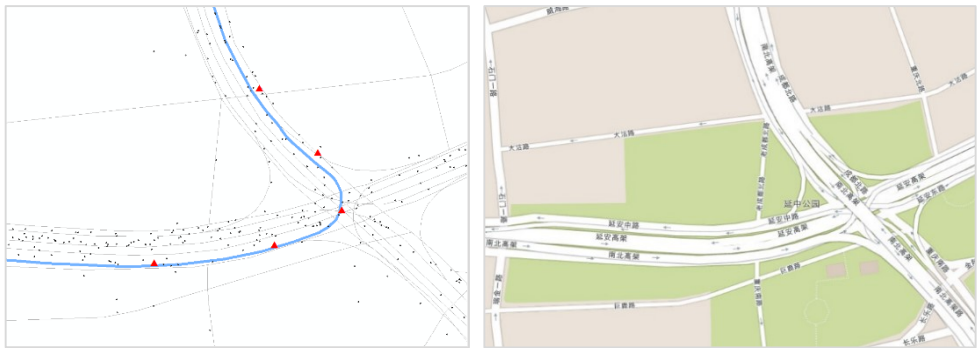

**Figure 1.** An example of GPS trajectory section in a dense road network. (Left) GPS observations depicted as red triangle along the true path in blue in the vector representation. The sampling interval is 10 seconds. (Right) Map view of roads in that area.

(Hunter et al., 2013) first introduced the CRFs to map matching in a real world project and achieved the best performance for a sampling interval of

60 seconds by building a complex model using 10 features. The CRFs shares with the HMM a similar factorization of probability computation, but is more flexible in using non-independent features. The success in practice and the flexible nature of CRFs has motivated us to incorporate complex features by leveraging existing modeling efforts in HMM-based methods. However, using a large number of features in the CRFs could cause over-fitting and increase computational expenses. To overcome this potential drawback, we investigate a $\ell_1$ regularized CRFs with the aim to set up a sparse model.

## 3. Map Matching with Conditional Random Fields

### 3.1. Conditional Random Fields

The Conditional Random Fields (CRFs) is an undirected graphical model used to compute probability of a possible label sequence conditioned on the observation sequence (Lafferty et al., 2001). The CRFs represents the conditional probability as the product of potential functions over cliques in the graph. These potential functions are computed in terms of feature functions of random variables in observation and label sequence. Let $Y = \{y_1, y_2, \ldots, y_T\}$ and $X = \{x_1, x_2, \ldots, x_T\}$ denote the label sequence and observation sequence. A CRFs formulates conditional probability of $Y$ given $X$ as:

$$P(Y|X) = \frac{1}{Z} \prod_q exp(\sum_k \omega_k f_k(Y_q, X_q))$$

where $\{f_k\}$ are the feature functions on any subset of the random variables in the sequence $Y_q \subset Y, X_q \subset X$ (note that $Y_q \bigcup X_q$ form the cliques in the graph), $\{\omega_k\}$ are the trained weights for each feature function, and $Z$ is a input-dependent normalization term over all possible state sequence:

$$Z = \sum_Y \prod_q exp(\sum_k \omega_k f_k(Y_q, X_q))$$

### 3.2. A CRFs Framework for Labeling GPS Trajectory

To apply the CRFs to map matching, we first define random variables to model the observation and label sequence. Let $X = \{x_1, .., x_N\}$ be GPS observation sequence, $Y = \{y_1, .., y_{2N-1}\}$ be the label sequence, $N$ be the length of the observation sequence and $t = 1..N$ be the position index in the sequence. We give the definition as follows:

- $x_t \in X$ is a variable representing GPS observation.

- $y_{2t-1} \in Y, t = 1..N$ is a random variable over point states $R^t = \{r_i^t,\}, i \in N_{2t-1}$ of observation $x_t$, where $R^t$ is a finite set of nearby roads of $x_t$ within a predefined distance.
- $y_{2t} \in Y, t = 1..N-1$ is a random variable over path states $P^{2t} = \{p_j^{2t}\}, j \in N_{2t}$, , where $P^{2t}$ contains all the feasible paths between road $r_i^t$ and road $r_j^{t+1}$. And $P^{2t}$ is also a finite set since vehicle can only travel a limited distance in a road network in specific time duration with speed limits.

Take abovementioned variables as the nodes, in which we call $\{x_t\}$ observation node, $\{y_{2t-1}\}$ point node and $\{y_{2t}\}$ path node. Then, we add edges between observation nodes and point node at each position $t$, while linking point nodes and path nodes sequentially. To be more concrete, we give a simplified example of chain structured CRFs for 3 GPS observations on road network in Figure 2. Note that applying different features to the model could result in a different topology between variable nodes and observation nodes.

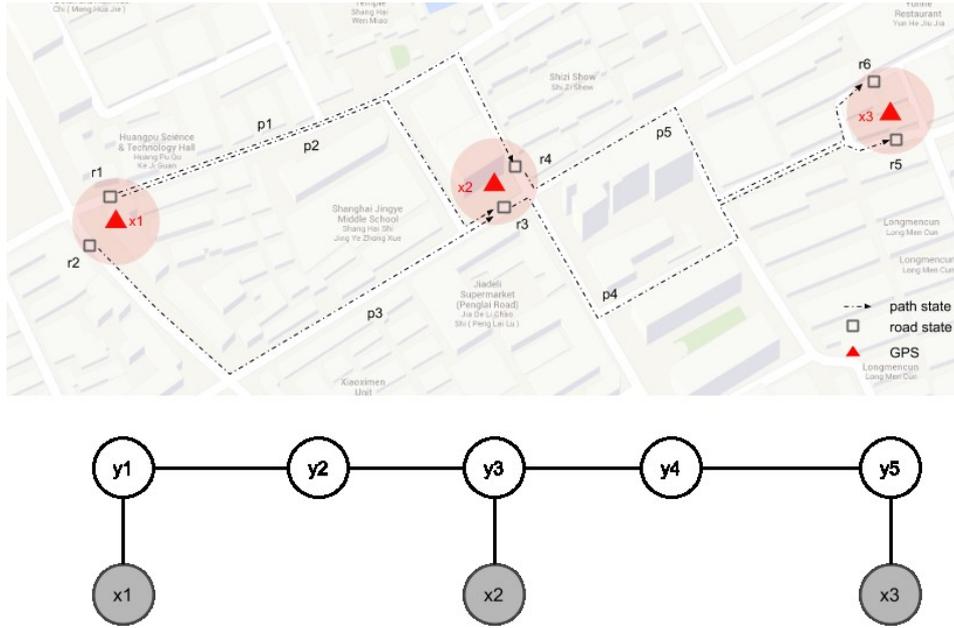

**Figure 2.** A chain-structured CRFs for 3 GPS observations. The map on top illustrates the simplified situation of identifying road states and path states given GPS observations in the road network. This requires 5 random variables, $y_1 : \{r_1, r_2\}, y_2 : \{p_1, p_2, p_3\}, y_3 : \{r_3, r_4\}$, $y_4 : \{p_4, p_5\}, y_5 : \{r_5, r_6\}$, to build the CRFs for map matching. Thus, nodes $y_1, y_3, y_5$ linking with observations (black circles) are point nodes while nodes $y_2, y_4$ are path nodes.

Then, a chain-structured CRFs for map matching is formulated as:

$$P(Y|X) = \frac{1}{Z} \prod_{t=1}^{N} exp(\sum_{k=1}^{K} \omega_k f_k(y_{2t-1}, x_t) + \sum_{s=1}^{S} \mu_s g_s(y_{2t}, y_{2t-1}, y_{2t+1}, X)) \quad (1)$$

where $\{f_k\}$ are feature functions defined on point nodes while $\{g_s\}$ are feature functions defined on path nodes (note that $g_s = 0$ when $t = N$), $\{\omega_k\}$ and $\{\mu_s\}$ are weights of the feature functions. These feature functions are designed to capture the characteristics of the actual label of point and path respectfully which we will discuss in detail in *Section 4*. And $Z$ is given as:

$$Z = \sum_{y \in Y} \prod_{t=1}^{N} exp(\sum_{k=1}^{K} \omega_k f_k(y_{2t-1}, x_t) + \sum_{s=1}^{S} \mu_s g_s(y_{2t}, y_{2t-1}, y_{2t+1}, X)) \quad (2)$$

The rationale of using path node to explicitly model transition between neighboring observations is that it allows the model to evaluate more than one path between two road states. This may avoid an early elimination of truth path state as most HMM-based methods are forced to use only one path, e.g. the shortest path in most cases. The modification is crucial especially for a low sampling rate of GPS trajectory which may lead to identification of many feasible paths. And our model differs from that by (Hunter et al., 2013) in the way that we encode the transition on the path node rather than on edges, which helps reduce the space complexity in later implementation.

### 3.3. Inference and Training on CRFs

Map matching can be casted as an inference on CRFs, which is to find the state sequence with the maximal probability conditioned on the observation sequences. For a general structured CRFs, the inference could become computationally intractable because the increasing length of the trajectory will exponentially enlarge the resulting state space. However, an exact solution can be obtained using dynamic programming algorithms such as Vite-bi on linear chain structure (Sutton, 2012).

The inference requires learned weights of the feature functions, which can be estimated by training CRFs with labeled data, namely GPS observation sequences are labeled against actual road sequences (also the road sequences in between). A common training scheme is to estimate the weights of the feature functions by maximizing the log likelihood function $log(P(Y|X; \omega, \mu))$, which yields

$$\ell(\omega,\mu) = log(P(Y|X;\omega,\mu))$$
$$= \sum_{t=1}^{N}(\sum_{k=1}^{K} \omega_k f_k(y_{2t-1}, x_t) + \sum_{s=1}^{S} \mu_s g_s(y_{2t}, y_{2t-1}, y_{2t+1}, X)) \quad (3)$$
$$- log(Z)$$

Since feature functions for point node and path node can be equally treated in the optimization, we rewrite the objective function for brevity as follows

$$\ell(\theta) = \sum_{t=1}^{N}(\sum_{m=1}^{K+S} \theta_m q_m) - log(Z) \quad (4)$$

Where
$$(\theta_m) = (\omega_1, \omega_2, ...; \mu_1, \mu_2, ...)$$
$$(q_m) = (f_1(\cdot), f_2(\cdot), ...; g_1(\cdot), g_2(\cdot), ...)$$

This yields a convex and differentiable objective function for which we can use unconstrained optimization method to find the global optimal solution. More specifically, a quasi-Newton method, BFGS, is used, which has been found successful in terms of efficiency for solving this objective function (Sha, Pereira, & Science, 2003).

## 4. Feature Selection in CRFs with $\ell_1$ Regularization

Often, to improve the classification result of the CRFs, more features should be used. However, this leads to a dilemma that using more features also increases the risk of over-fitting. Therefore, it has been a long-term endeavor in machine learning community to study feature selection with the aim to find the most relevant feature subset to build a compact and interpretable model (Ng, 1998). This involves two tasks, feature induction and feature selection. We discuss them in the following sections in the context of map matching.

### 4.1. Feature Induction and Parameter Tying

In our chain structured model, two types of feature functions are used, namely point features and path features. Both features can be designed in an either manual or automated fashion to capture the characteristics of truth states. We employ both strategies for the feature induction.

Hand-crafted features are extracted from HMM-based map matching methods (Goh et al., 2012; Hummel, 2006; Krumm et al., 2007; Lou et al., 2009; Newson & Krumm, 2009; Yuan et al., 2010). As HMM shares with the CRFs a similar structure, it is straightforward to derive point feature and path feature from the emission probability and the transition probability. Note that most of these probabilities follow the assumption of Gaussian distribution. Since the CRFs uses exponential parameterization, only the informative power terms in the formulation of HMM-based methods are needed. The detailed formulation is given in *Section 5*.

Another observation for map matching is that routing on road network is a dynamic process. Driver behaviors could vary a lot in different spatiotemporal contexts. For example, paths along main roads could be avoided to dodge heavy traffic in rush hours while they are taken during the night. In order to feed the model with this kind of contextual information, we employ two feature templates coding road class and temporal information for point and path nodes:

$$f = \mathbf{I}(y_{2t-1} = r_i^t)\mathbf{I}(r_i^t = \text{class}_u)h(y_{2t-1}, X) \tag{5}$$

$$g = \mathbf{I}(y_{2t} = p_j^{2t})\mathbf{I}(t \text{ in period}_v)h(y_{2t}, X) \tag{6}$$

Where **I** is indicator function, it yields 1 when the given condition holds, 0 otherwise, $h(y, X)$ could be any feature functions revealing contextual characteristics of the ground truths (e.g., varying travel speed in different periods during the day), $\{\text{class}_u\}$ and $\{\text{period}_v\}$ are sets of road classes and time periods in hour during the day. The road class information is extracted from OpenStreetMap (OSM) and timespans are divided using predefine time interval.

Feature template could result in a large number of features. For example, $M$ feature functions with $N$ would need $MN$ parameters. This imposes a heavy computational load for the later training as $N$ is usually very large for the low sampling rate trajectory. However, the CRFs for map matching serves as a structured classifier to perform binary classification of truth/false state for both point node and path node. And there is no need to assign a unique parameter for individual point states and path states at each position in the chain. Therefore, only $M$ parameters are needed in the model.

### 4.2. Training CRFs with $\ell_1$ Regularization

To achieve a better classification performance in map matching, a large number of features are used in the CRFs. This yields a lower error rate on training data while raising the risk of high generalization error on test data. A common technique to tackle this problem is to add a penalty term to the objective function which penalizes learning large weights of feature functions in training. In this section, we discuss two kinds of regularization techniques, $\ell_2$ regularization and $\ell_1$ regularization, and explain how to perform the feature selection with the latter one.

1. $\ell_2$ Regularization

$\ell_2$ regularized CRFs adds a negative quadratic term to the objective, which tends to keep all the weights small enough but non-zero in training. This yields to solve:

$$\max_\theta \ell(\theta) - \lambda_2 (\sum_m \theta_m) \qquad (7)$$

Here $\lambda_2 > 0$ is a hyper parameter that controls the amount of the penalty: the larger the value of $\lambda_2$, the greater the amount of penalty and 0 for no penalty. Since the penalty term is differentiable with respect to parameters of the model, the objective remains convex and differentiable. Therefore the optimization method used to train non-regularized CRFs can also be applied here.

2. $\ell_1$ Regularization

Another regularization technique, $\ell_1$ regularization, adds absolute term to the objective, which tends to reduce the weights to exactly zero in training. This yields to solve:

$$\max_\theta \ell(\theta) - \lambda_1 \sum_m |\theta_m| \qquad (8)$$

where $\lambda_1 \geq 0$ again is used to tune the amount of penalty. The objective also remains convex while become non-differentiable at $\theta_m = 0$, which requires extra treatment to solve this optimization problem.

Having the advantage of producing a sparse model (having many parameter set to 0), optimizing $\ell_1$ regularization has invoked a lot of interest in machine learning community. A variety of optimization methods are proposed to solve the problem. Since the convexity of $\ell_1$-regularized objective ensures the finding of a unique optimal solution, those methods can be distinguished by how they handle non-differentiability of the objective function. Therefore, we mainly consider the efficiency in terms of running time while choosing optimization algorithms. Some comprehensive experimental reviews have been reported in (Schmidt, Fung, & Rosaless, 2009; Schmidt, 2010), which stimulated our interest in the Projected Scaled Sub-Gradient (PSS) methods for its fast convergence rate and consistent performance across different types of data set. We also find it more successful on GPS trajectory data.

Still, we have to choose the hyper parameters $\lambda_1$ and $\lambda_2$ which are difficult to determine in advance. As for $\lambda_1$, we tune the hyper parameters by evaluating the resulting error rates using a geometric sequence of decreasing from $\lambda_{max}$ to 0, where $\lambda_{max}$ is large enough to reduce all weights to zero. The justification of using a geometric sequence is that the target value is close to 0 and more trials are needed to approach it. And we use the same hyper parameter for $\ell_2$ for comparison.

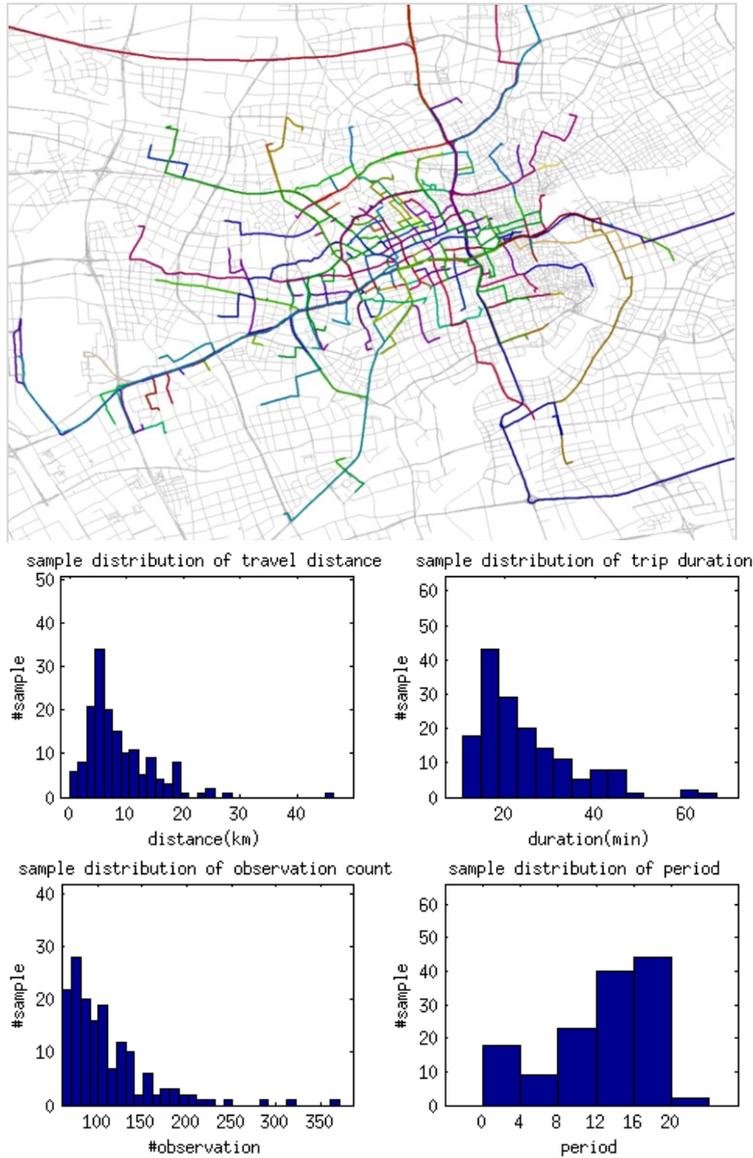

**Figure 3.** (Top) The spatial distribution of GPS trajectories in ST100. (Bottom) The statistics of sample trajectory in ST100: travel distance (upper left), trip duration (upper right), observation count (bottom left) and daytime period in hour (bottom right).

## 5. Experiment

We build a compact CRFs for map matching of low sample rate GPS trajectories by training the model with $\ell_1$-regularization. To examine the efficiency of $\ell_1$-regularization, we test our methods on a GPS trajectory sample dataset ST100 to compare its error rates of map matching with the CRFs trained with a common $\ell_2$-norm. In following sections, we first introduce the sample dataset, and then describe features for map matching. In the end, the results are discussed.

### 5.1. Experiment Setup

The sample dataset ST100 records GPS trajectories of 70 taxis from one day across the downtown area in Shanghai, China. It involves 124 trajectories in total and 13767 GPS observations covering an overall length of 788 km after eliminating some erroneous trajectories, e.g. extremely short trips and trips losing long distance GPS observations. Spatial distribution of the trajectories in ST100 and statistics of sample trajectories are demonstrated in *Figure 3*.

As the data source doesn't provide the ground truth labels, we have to manually label them on the reference road network. We recruited 2 volunteers with driving experiences in China to trace the trajectories on the map using OSRM, a web-based interactive routing application using road data from OSM. For routing exceptions like U-turns, it requires manual post-processing individually.

In order to test the consistency of our models at different sampling rates, we degrade ST100 to three datasets with 60, 90 and 120-second interval accordingly using an even sampling strategy. For each degraded dataset, we split it into a training set and a test set with a ratio of 7:3. A portion of training set is used as hold out data to tune the hyper parameters. These settings are applied to both $\ell_2$ and $\ell_1$ regularization.

### 5.2. Features for Map Matching

Here we give details of the features used in the CRFs for all experiments. There are in total 61 features in the model. 9 of them are derived from HMM-based methods in the literatures and most of the rest are designed to reveal the road usage pattern and temporal behavior of the drivers. For brevity, we omit the dummy term $\mathbf{I}(y_{2t-1} = r_i^t)$ for point node feature and $\mathbf{I}(y_{2t} = p_m^{2t})$ for path node features. And we set time interval to 4 hours for all temporal features.

| Feature | Description | Node type |
| --- | --- | --- |
| $\mathrm{dist}(x_t, r_i^t)$ | *GPS distance error* between GPS observation and road candidate. | point |
| $\mathrm{aziumth}(x_t, r_i^t)$ | *angular difference* between vehicle's heading direction and the road direction. | point |
| $(v(x_t) - v(r_i^t))/v(r_i^t)$ | *speed difference ratio* between vehicle and speed limits of the road | point |
| $\mathbf{I}(t \text{ in period}_v)(v(x_t) - v(r_i^t))/v(r_i^t)$ | *temporal speed difference ratio* | point |
| $\mathbf{I}(r_i^t = \text{class}_u)\mathbf{I}(\text{taxi in servie})$ | *road usage* indicate how often a certain class of roads is used when the taxi is with passenger. | point |
| $\mathbf{I}(r_i^t = \text{class}_u)\mathbf{I}(t = 1 or N)$ | *IO* feature indicates the road usage when the taxi picks up or drops off passengers. | point |
| $\mathrm{length}(p_j^{2t})$ | *length* of the path | path |
| $\mathrm{t_{min}}(p_j^{2t})$ | *minimum travel time* on the path, using speed limits of the roads and the time interval between GPS observations | path |
| $\overline{v}(p_j^{2t})$ | *maximum average speed* is the average speed limits of roads in the path | path |
| $\mathrm{dist}(x_t, x_{t+1})/\mathrm{length}(p_j^{2t})$ | *length ratio* of distance between GPS observations to the path's length | path |
| $\cos(v(p_j^{2t}), \overline{v}(p_j^{2t}))$ | *cosine distance* between the speed limits of the roads in path and the overall average speed on the path | path |
| $\mathrm{length}(p_j^{2t}) - \mathrm{dist}(x_t, x_{t+1})$ | *length difference* between the path and the distance between GPS observations | path |
| $\mathrm{t_{min}}(p_j^{2t}) - t(x_t, x_{t+1})$ | *Time difference* between the estimated the minimum travel time and the actual travel time. | path |
| $\mathrm{classmod}(p_j^{2t})$ | *road class changes* in the path | path |
| $\mathbf{I}(t \text{ in period}_v)(\mathrm{length}(p_j^{2t}) - \mathrm{dist}(x_t, x_{t+1}))$ | *temporal length difference* of the path | path |
| $\mathbf{I}(\text{taxi in service})\mathrm{classmod}(p_j^{2t})$ | *temporal road class changes* | path |

**Table 1.** Features used in the model.

These feature data are generated from the ST100 and the OSM road network on Postgreql with PostGIS and pgRouting. The spatial extensions are used to perform spatial queries and graph search to identify road states and path states. Before the data are fed to the CRFs, we rescale the features to the range $[0, 1]$ so as to avoid a dominant impact of some features with large values on the model.

### 5.3. Matching Results of Low Sampling Rate GPS Trajectory

We tested our model with two different regularization on three sets of low sampling rate GPS trajectories from sample dataset ST100 and compared the error rate of point and path separately. Though point nodes and path nodes have mutual impacts on each other in the chain structure of the CRFs, labeling path is usually more difficult than labeling points. Therefore, we chose to evaluate the performance on different nodes individually.

The matching results on three sampled datasets are summarized in *Table 1*. It shows that the error rate increases as the sampling interval grows in which the path error rates deteriorate faster than the point error rates. The reasons that path matching is more challenging are 1) path features fail to discriminate the actual paths among the huge numbers of the path candidates; 2) the routing preference might not be consistent across the trajectories. With regard to the two regularizations, training CRFs with $\ell_1$-norm managed to reduce more than half of the features that are necessary with $\ell_2$ while achieving an average of 10% reduction on the error rates. Meanwhile, we also compare $\ell_1$-regularized CRFs with *MaxLL-complex* from (Hunter et al., 2013). Both achieve only the same accuracy performance on the test data of 120s interval. However, our method is more flexible because we give alternative choices of features, which could be helpful when desired features are not available in the data (e.g., POIs of traffic lights are not available in the test area in OSM).

| Intervals | regularizer | feature number | point error rate | path error rate |
|---|---|---|---|---|
| 60 | $\ell_2$ | 44 | .228 | .299 |
|  | $\ell_1$ | 18 | .153 | .194 |
| 90 | $\ell_2$ | 43 | .235 | .304 |
|  | $\ell_1$ | 20 | .146 | .197 |
| 120 | $\ell_2$ | 43 | .255 | .339 |
|  | $\ell_1$ | 17 | .166 | .234 |

**Table 1.** Error rates of CRFs with different regularizers for map matching.

We examined the effectiveness of feature selection by tuning hyper parameter on hold out dataset with 120 second sampling intervals. As shown in *Figure 5*, features are gradually added to the model when decreasing λ. And the error rate is reduced as more informative features are used. The improvement stopped at some tipping point where adding more features may cause the model to over fit the training data.

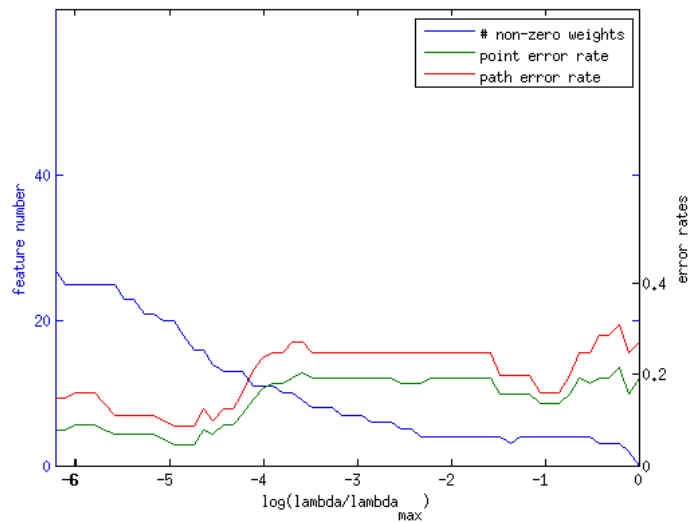

**Figure 5.** The impact of tuning $\lambda$ on the number of selected features (non-zero weights) and error rates.

In the end, we evaluate all the selected features (only those have non-zero weights in all three tests) in *Figure 6*. The weights vary dramatically across the selected features, in which 1~10 are for point features and 11~13 are for path features. The weight magnitude shows the relevance degree of the feature to the map matching task, while the sign of weights indicate how the features are related. For example, *GPS distance error* getting a negative weight means that the states are more likely to be true if the its *GPS distance error* are smaller. Among all the features, *GPS distance error* (1), *length difference* (12) and *road class changes* (13) are the most relevant ones. The negative weight of *road class change* indicates that the drivers prefer to stay on the roads of the same class. The weights of feature *road usage* (3~8) show that the road usages are unbalanced between the taxis in service (ru-1) and not (ru-0). Two classes of roads are selected in the feature *IO* (9~10) with relatively large negative weights indicating that taxis may barely pick up and drop off passengers there.

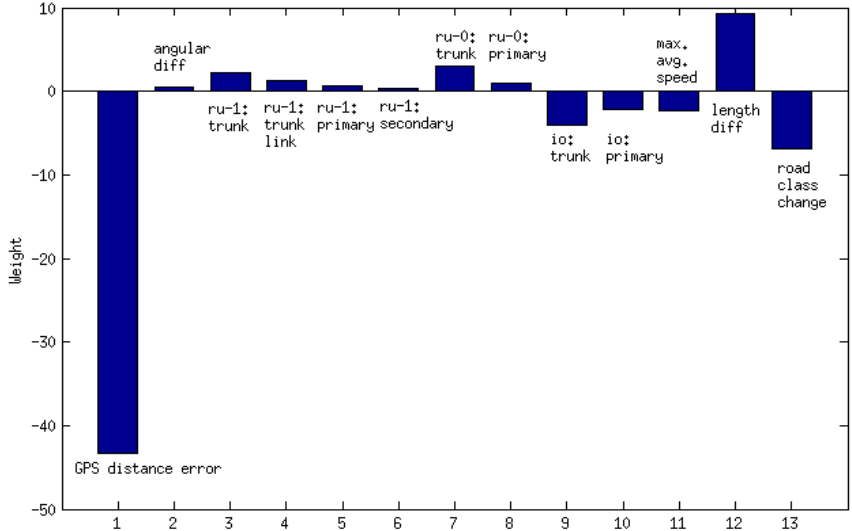

**Figure 6.** The weights of selected features learned from the data.

## 6. Conclusion and Future Work

By inducing complex and non-independent features, we explored the use of CRFs for map matching GPS trajectory at a low sample rate. Rather than using a common $\ell_2$-regularization, we train the CRFs with a $\ell_1$-norm to yield a sparse model which requires less computation cost to perform the map matching. To verify the model, we build a sample dataset, ST100, from Shanghai Taxi FCD. Experiments on ST100 have shown the effectiveness of $\ell_1$-regularization on both feature selection and matching accuracy. The result of feature selection can provide a guidance to build a compact model and meanwhile reveals to a certain extent the pattern of road usage in the urban road network of our study area.

In the future work, we intend to improve the method from following perspectives: 1) induce context-aware features to capture the spatial variance of routing decisions on urban road network for map matching; 2) study efficient training method for the CRFs with a larger feature set.

## 7. Acknowledgement

We would like to thank Dr.-Ing Hongchao Fan and Prof. Chun Liu for sharing with us the Shanghai Taxi FCD dataset, and to Oliver Maksymiuk for the helpful discussion. The first author is supported by China Scholarship Council (CSC).